\documentclass[subscriptcorrection,upint,varvw,barcolor=Goldenrod3,mathalfa=cal=euler,balance,hyphenate,french,nocopyright,nolists]{FNODE} %

\usepackage{hyperref} 
\usepackage{hyperxmp} 
\usepackage[x11names,svgnames]{xcolor}
\usepackage{tikz}
\usetikzlibrary{arrows}      
\usetikzlibrary{arrows.meta} 
\usepackage{float}
\usetikzlibrary{shapes.geometric, arrows.meta, positioning, fit, backgrounds}
\usetikzlibrary{decorations.pathmorphing}
\usepackage{fancyhdr}

\hypersetup{%
	pdfauthor={Mohammed Ashraf},                       		   	
	pdftitle={Frequency-Domain Neural ODEs for Modeling Non-Linear Dynamical Systems},                  	
	pdfkeywords={Control Theory, Dynamical Systems, Neural Networks, Continuous-time models, Discrete-time models, Fast Fourier Transform, Data-Driven Models},
	pdfsubject = {System prediction using NeuralODEs},			
	pdflicenseurl={https://ctan.org/pkg/asmejour},
}

\hypersetup{
    pdfa=true,
    pdfapart=2,
    pdfaconformance=u
}

\PaperYear{}
                   
\begin{document}


\SetAuthorBlock{Mohammed Ashraf}{Department of Mechatronics Engineering,\\
   German University in Cairo,\\
   Cairo, Egypt \\
   email: mohammed.abdelrehim@guc.edu.eg
   } 


\SetAuthorBlock{Ayman A. El-Badawy}{%
Department of Mechatronics Engineering,\\
German University in Cairo, \\
Cairo, Egypt \\
email: ayman.elbadawy@guc.edu.eg
}


\title{Frequency-Domain Neural ODEs for Modeling Non-Linear Dynamical Systems}


\begin{abstract}
\textbf{Abstract:} Standard continuous-depth models, such as Neural Ordinary Differential Equations (NODEs), offer significant advantages in modeling physical systems by learning continuous vector fields rather than discrete temporal steps. However, when applied to complex dynamical systems, standard NODEs frequently struggle with highly nonlinear dynamics. This paper investigates the Frequency-domain Neural ODE (FNODE), an architecture that projects continuous temporal dynamics into the frequency domain using the Fast Fourier Transform (FFT). By operating in the frequency domain, the model provides better generalization to the dynamical system. The architecture is empirically evaluated against discrete models, specifically Gated Recurrent Units (GRUs) and Long Short-Term Memory (LSTMs), and other continuous-depth variants, including Augmented Neural ODE (ANODE), across four distinct dynamical systems: the Lotka-Volterra model, the forced Duffing oscillator, the Van der Pol oscillator, and the Lorenz system. To rigorously assess generalization and robustness, curriculum and ensemble learning are used to evaluate the model’s convergence by estimating confidence intervals across different ensemble models. The empirical results demonstrate that the FNODE architecture achieves better generalization while exhibiting remarkable convergence stability.
\end{abstract}

\date{Version \versionno, \today}

\maketitle 


\section{Introduction}   

The domain of using data-driven techniques for identifying systems from raw data has grown significantly over the past decade. The aim of system identification is to deduce a mathematical relationship between the past input-output data and the future state of a system, which requires a high level of intervention and expertise \cite{sjoberg1994neural}. The recent rise in deep learning models and computational power has improved the performance of data-driven approaches. The application of data-driven techniques is not only a modern alternative but also a dominant approach to uncovering the underlying nature of complex systems by extracting relevant information from observational data \cite{HORNIK1989359, mienye2024comprehensive, LJUNG20201175, pillonetto2024deep}.

Although classical mathematical modeling has shown great promise in past decades, it is known to show signs of limitation when applied to complex systems with insufficient modeling. Generally, classical modeling methodologies depend on a set of rigid equations based on a complete understanding of the underlying physics and have shown great promise in terms of sample efficiency, along with strong theoretical guarantees \cite{lu2019DeepXDEAD}. These modeling techniques often depend on strong assumptions, not necessarily true in real-world scenarios. In cases where extreme nonlinearities, chaotic transients, or environmental interactions are not known, such as in cases involving turbulent fluid flow or biological processes, an accurate analytical representation is mathematically impractical to derive \cite{weinan2017proposal}. Additionally, the computational cost of accurately solving such a complex representation using classical methodologies, such as numerical integration, requires extensive domain knowledge to balance model complexity with computational speed \cite{YAMADA2023102, ogunmolu2016nonlinear}. 

As a solution to these complexities in classical mathematics, the paradigm of scientific machine learning (SciML) has been established as a viable solution. SciML does not aim to replace the need for physics by depending on machine learning techniques; instead, it attempts to merge these two concepts by discovering the underlying equations through data, which remain impossible to derive \cite{cuomo2022scientific}. This allows for the development of surrogate models that can mimic complex phenomena with much less human intervention compared to traditional approaches. SciML has addressed the modelling bottleneck by directly incorporating physical constraints, enabling the discovery of systems that have traditionally been considered "too noisy" or "too stiff" for traditional solution methods \cite{cuomo2022scientific}. 

The scientific machine learning paradigm is a part of the universal function approximator capabilities of neural networks (NNs) and deep learning (DL). At a fundamental level, a neural network can be viewed as a mathematical framework that can efficiently map complex input spaces of high dimensionality onto latent output manifolds of comparable precision \cite{popescu2009multilayer, lecun2015deep}. Within the context of dynamical systems, neural networks have been utilized to approximate the “right-hand side” of a differential equation, which specifies the vector field that dictates the state’s evolution in time. Since deep neural networks can more efficiently circumvent the “curse of dimensionality” compared to shallow networks, they can more efficiently model the high-dimensional latent spaces that are ubiquitous in physical systems. These capabilities in modeling a continuous function from data allow for the identification of nonlinear dynamics without the need for a prior specification of specific functions or physical constants that dictate the motion \cite{poggio2017why}. 

\section{Literature Review}

With the increasing need to model and predict time-series data, there was a shift in focus towards models that were specifically designed to handle sequential data. Although feedforward models were highly efficient, they did not have an inbuilt capacity to account for the sequence and history of observations. To resolve this, Recurrent Neural Networks (RNNs) were proposed, which have an inbuilt hidden state that can evolve over time as more observations in the sequence are processed \cite{mienyernn2024, COSSU2021607}. Although RNNs were highly efficient, they were unable to learn long-term dependencies within a sequence, leading to the vanishing gradient problem. This problem was later addressed by the Long Short-Term Memory (LSTM) network, which has an inbuilt complex gating mechanism to determine which information should be stored or discarded over long periods of time. This became the standard approach to time series forecasting, which could reliably capture the memory of a physical system as it moves through different states \cite{hochreiterlstm1997, lindemannlstm2020, ALSELWI2024102068}. Gated Recurrent Unit (GRU), a variant of the standard RNN, was specifically designed to overcome the vanishing gradient problem and capture long-term temporal dependencies in sequential data. GRUs regulate the flow of information using a specialised internal gating mechanism. Unlike LSTM, the GRU simplifies the network by merging the internal memory and utilising only two dynamic gates \cite{GRU2021}.

Even with their successes in areas with sequential and temporal data, RNNs and LSTMs have a basic drawback in dealing with physical time series, where their ability to handle time in a discrete manner is a critical drawback. In RNNs and LSTMs, it is generally considered that the change from a certain hidden state to the following hidden state takes place at regular intervals of time, which is rarely found in real-world engineering scenarios where sampling may not always take place at regular intervals or the presence of communication delays in measurement systems \cite{caobias2021}. Since these models are restricted to a particular sampling rate with which they were trained, they rarely generalize to a different sampling rate in the testing environment, which is a significant drawback of these models. By dealing with time in a discrete manner, these models struggle with continuous changes in the underlying physics, where their predictions are found to be inconsistent over long periods of time due to the spectral bias towards low frequency of deep neural network learning methods in learning multiple-frequency solutions \cite{XU2025113905}. 

In order to overcome these limitations, the idea of continuous depth models has been adopted. In this approach, instead of modeling the sequence of layer transitions, the hidden state is represented as a continuous function with respect to time. This approach enables the neural network to approximate the derivative of the system's state, effectively transforming the network into an equation that can be evaluated at any arbitrary time point \cite{Hasani2022}.

Before the advent of continuous-depth neural networks, various alternative methods were used to effectively incorporate physical knowledge into the data-driven modeling approach. One such important method is the use of the Sparse Identification of Nonlinear Dynamics (SINDy) approach, where sparse regression is used to determine the simplest set of governing equations from noisy data \cite{brunton2016sindy}. Furthermore, Gelß et al. proposed multidimensional approximation of nonlinear dynamical systems (MANDy), which combines data-driven methods with tensor network decompositions to reduce the computational costs and storage consumption for high-dimensional datasets to avoid the curse of dimensionality \cite{MANDy}. Another important innovation is the Physics-Informed Neural Network (PINN) approach, where existing physical laws such as the set of partial differential equations are incorporated as constraints within the loss function \cite{karniadakis2021physics, robinson2022physics}. This approach enables networks to learn from significantly less data compared to the traditional black-box approach. Despite the fact that these methods are important tools for discovering physical laws, they require significant prior knowledge about the underlying structure of the system, which is an important challenge for modeling highly complex systems \cite{wangpinn2022}. 

The most significant change in continuous-time modeling was developed by Chen et al. with the introduction of Neural Ordinary Differential Equations (NODEs). This framework changes the understanding of deep neural network architecture by viewing a series of hidden layers as a discretization of a continuous-time process \cite{chen2018neural}. In a Residual Network (ResNet), a series of layers performs a discrete process on the current state \cite{he2016resnet}. NODEs, on the other hand, examine this in the limit as the step size goes to zero, resulting in a continuous vector field represented by a neural network. By replacing a standard series of layers with a black-box differential equation solver, the forward pass of a neural network can be viewed as an initial value problem (IVP). Moving from a discrete system of transitions to a continuous geometric system, this architecture is able to represent a system without a fixed depth, effectively a network of “infinite” depth \cite{kidger2022node, massaroli2020node, ai6050105}. 
 
Another advantage to NODE architecture is its unique combination of adaptive computation and constant memory efficiency. By applying the adjoint sensitivity method for calculating gradients, NODEs allow for the use of the backpropagation of derivatives through the ODE solver without the need to store intermediate activations in memory. This is particularly useful for the training of very complex models on platforms that have limited memory resources available \cite{baydin2015AutomaticDI, amores2025autodiff}. Additionally, the use of adaptive numerical solvers allows for the dynamic adaptation of the evaluation strategy depending on the complexity of the input trajectory. During the inference phase of the network, there is the option to balance numerical accuracy against computational speed by adjusting the error tolerance of the solver. This is useful for the deployment in real-time systems \cite{worsham2025guide, lee2021node, gaurav2025neural}. 

While the theoretical foundation of Neural Ordinary Differential Equations (NODEs) is attractive, a number of limitations have been identified, particularly when applied to complex dynamical systems. The first limitation is a topological one. Because of the inability of trajectories to cross in a, the NODE architecture is mathematically forced to learn continuous invertible mappings. This limitation prevents the model from learning simple mappings or complex dynamics where trajectories appear to cross or intersect in a lower-dimensional embedding \cite{rodriguez2022LyaNet}. Moreover, they are inclined to be numerically stiff, particularly when modeling complex dynamics where trajectories are chaotic and highly oscillatory. In such systems, the underlying vector field is extremely sensitive to perturbations, and the adaptive solver is forced to take extremely small time steps to maintain accuracy, leading to a computationally costly solution \cite{caldana2025}. 

In order to overcome these drawbacks, several variants of the NODE have been developed to improve the original framework. To overcome topological constraints, Augmented Neural ODEs (ANODEs) have been proposed by Dupont et al. (2019), which augment the state space by introducing additional dimensions. This provides the flexibility required for the model to learn complex mappings that cannot be learned by the original input space, thus improving generalization and training stability \cite{dupont2019AugmentedNO}. Further research into NODEs has also focused on improving the computational efficiency of the solver itself. The Spectral Discretization of Neural ODEs (SNODE), proposed by Quaglino et al. (2020), replaces explicit integration in the time domain by a spectral element method that represents the dynamics by a series of Legendre polynomials. The coefficients of the Legendre polynomial are optimized instead of discretizing the time domain and integrating over it. The spectral discretization method provides a significant speedup and allows for time-parallel training \cite{quaglino2019SNODESD}. Although these models have improved the robustness and flexibility of NODEs, these models are also limited to the time domain and are prone to local truncation errors in long-term forecasting \cite{chen2023node, linot2022node}. 

Despite the significant variability of these refinements—from state space augmentation in ANODEs to polynomial series approximations in SNODEs—the overwhelming majority of contemporary forms of Neural Drdinary differential Equations possess a basic architectural constraint that cannot be overcome: namely, these models remain fundamentally tied to the time domain. Whether the purpose of a new refinement is to improve topological expressivity or computational speed, the underlying imperative of every contemporary form of Neural Ordinary Differential Equation is the sequential integration of localized geometric tangents \cite{goyal2023neural}. This basic architectural constraint ensures that the predictive accuracy of these models will forever be tied to the quality of the neural vector field and the numerical engine. 

However, a more conceptual and underlying shortcoming of these models is the inherent "spectral bias" of these models, a phenomenon inherent to the nature of gradient-based optimization of deep neural networks. Indeed, by the standards of the theoretical framework of the Neural Tangent Kernel (NTK), over-parameterized models are shown to have an inductive bias towards learning low-frequency and smooth components of a signal while remaining essentially "blind" to high-frequency transients and multi-scale oscillations \cite{turan2021MultipleSF}. When faced with the problem of modeling the dynamics of systems such as energy storage systems, these models will naturally "flatten out" the high-frequency oscillations in an effort to find a smoother and lower energy solution \cite{gao2023neural}. This leads to an approximated vector field that is topologically incorrect and fails to capture the sharp phase transitions and energy bursts required to maintain a stable attractor and subsequently leads to a collapse of the predicted trajectory \cite{dang2023CDNODE, ghosh2020steer}.

Neural Ordinary Differential Equations (NODEs) and their augmented variants have demonstrated success in identifying continuous-time dynamics; the literature highlights a persistent vulnerability when applied to highly oscillatory or chaotic regimes. To bridge this gap, this paper proposes a novel architecture: Frequency Neural ODE (FNODE) architecture. Rather than stepping through physical time, the proposed method projects the full temporal window into the frequency domain, mitigating spectral leakage, and integrates the spectral coefficients over a continuous depth. By operating on the global frequency rather than local spatial coordinates, this approach aims to preserve the exact topological structure of complex limit cycles and chaotic attractors without the degradation inherent to traditional time-domain solvers.

\section{Model Architectures}

\subsection{Residual Networks (ResNets)}

Residual Networks (ResNets) were initially introduced to address the degradation problem encountered when training very deep neural architectures, commonly known as the vanishing gradient problem. As networks become deeper, standard feed-forward architectures often suffer from vanishing or exploding gradients, limiting their representational capacity. ResNets mitigate this bottleneck by redefining the learning objective of the hidden layers \cite{kidger2022node}. 

Let $\mathbf{z}_k$ represent the hidden state representation at the $k$-th discrete layer. A standard feed-forward network computes the next state as $\mathbf{z}_{k+1} = f_\theta(\mathbf{z}_k)$, where $f_\theta$ denotes the non-linear transformation parameterized by weights $\theta$. A ResNet introduces skip connections (identity mappings) that bypass the non-linear transformation block. The state update sequence is defined as:
\begin{equation}
    \mathbf{z}_{k+1} = \mathbf{z}_k + f_\theta(\mathbf{z}_k)
    \label{eq:resnet}
\end{equation}

This formulation ensures that the identity mapping can pass the gradient backward through the network. Moreover, Equation \ref{eq:resnet} exhibits strict mathematical equivalence with the explicit Euler method for numerical integration. By interpreting the sequence of discrete hidden layers as iterative time steps evaluating a dynamical system with a fixed step size $\Delta t$, the ResNet architecture represents a discretized, step-wise approximation of a continuous trajectory.

\subsection{Time-Domain Neural Ordinary Differential Equations (NODE)}

\begin{figure*}[t]
\centering
\resizebox{\textwidth}{!}{%
\begin{tikzpicture}[>=Stealth, font=\sffamily,
    dot/.style={circle, fill=black, inner sep=1.2pt},
    reddot/.style={circle, fill=red!80!black, inner sep=1.2pt},
    jump/.style={->, dashed, blue!70!black, thick},
    axisline/.style={->, thick, black!80}
]


\begin{scope}[shift={(0, 0)}]
    \node[font=\bfseries, text=blue!80!black] at (2.5, 5.7) {Residual Network};
    
    \draw[thick, black] (0, 0) rectangle (5, 5);
    
    \foreach \y in {0, 1, 2, 3, 4, 5} {
        \node[left=4pt, font=\scriptsize] at (0, \y) {\y};
    }
    \node[rotate=90, font=\scriptsize] at (-0.8, 2.5) {Depth};
    
    \node[below=6pt, font=\scriptsize] at (0.5, 0) {$-5$};
    \node[below=6pt, font=\scriptsize] at (2.5, 0) {$0$};
    \node[below=6pt, font=\scriptsize] at (4.5, 0) {$5$};
    \node[below=18pt, font=\scriptsize] at (2.5, 0) {Input/Hidden/Output};
    
    \begin{scope}
        \clip (0, 0) rectangle (5, 5);
        
        \foreach \x in {0.4, 1.2, 2.0, 2.8, 3.6, 4.4} {
            \foreach \y in {0.5, 1.5, 2.5, 3.5, 4.5} {
                \pgfmathtruncatemacro{\pct}{20*\x}
                \pgfmathsetmacro{\vx}{0.18 * (\x - 2.5) * (\y*0.4 + 0.4)}
                \pgfmathsetmacro{\vy}{0.4}
                \draw[->, color=red!\pct!blue, opacity=0.45, thick] (\x, \y) -- ++(\vx, \vy);
            }
        }
        
        \foreach \startx in {1.9, 2.15, 2.35, 2.65, 2.85, 3.1} {
            \draw[thick, black] (\startx, 0) 
                -- ({2.5 + (\startx-2.5)*(1 + 0.18*1*1)}, 1) node[dot] {}
                -- ({2.5 + (\startx-2.5)*(1 + 0.18*2*2)}, 2) node[dot] {}
                -- ({2.5 + (\startx-2.5)*(1 + 0.18*3*3)}, 3) node[dot] {}
                -- ({2.5 + (\startx-2.5)*(1 + 0.18*4*4)}, 4) node[dot] {}
                -- ({2.5 + (\startx-2.5)*(1 + 0.18*5*5)}, 5) node[dot] {};
            \node[dot] at (\startx, 0) {};
        }
    \end{scope}
\end{scope}

\begin{scope}[shift={(6, 0)}]
    \node[font=\bfseries, text=blue!80!black] at (2.5, 5.7) {ODE Network};
    
    \draw[thick, black] (0, 0) rectangle (5, 5);
    
    \foreach \y in {0, 1, 2, 3, 4, 5} {
        \node[left=4pt, font=\scriptsize] at (0, \y) {\y};
    }
    \node[rotate=90, font=\scriptsize] at (-0.8, 2.5) {Depth};
    
    \node[below=6pt, font=\scriptsize] at (0.5, 0) {$-5$};
    \node[below=6pt, font=\scriptsize] at (2.5, 0) {$0$};
    \node[below=6pt, font=\scriptsize] at (4.5, 0) {$5$};
    \node[below=18pt, font=\scriptsize] at (2.5, 0) {Input/Hidden/Output};
    
    \begin{scope}
        \clip (0, 0) rectangle (5, 5);
        
        \foreach \x in {0.25, 0.75, 1.25, 1.75, 2.25, 2.75, 3.25, 3.75, 4.25, 4.75} {
            \foreach \y in {0.25, 0.75, 1.25, 1.75, 2.25, 2.75, 3.25, 3.75, 4.25, 4.75} {
                \pgfmathtruncatemacro{\pct}{20*\x}
                \pgfmathsetmacro{\vx}{0.18 * (\x - 2.5) * (\y*0.4 + 0.4)}
                \pgfmathsetmacro{\vy}{0.35}
                \draw[->, color=red!\pct!blue, opacity=0.45, thick] (\x, \y) -- ++(\vx, \vy);
            }
        }
        
        \foreach \startx in {1.9, 2.15, 2.35, 2.65, 2.85, 3.1} {
            \draw[thick, black, smooth, samples=50, domain=0:5] 
                plot ({2.5 + (\startx - 2.5)*(1 + 0.18*\x*\x)}, \x);
                
            \node[dot] at (\startx, 0) {};
            \node[dot] at ({2.5 + (\startx - 2.5)*(1 + 0.18*0.6*0.6)}, 0.6) {};
            \node[dot] at ({2.5 + (\startx - 2.5)*(1 + 0.18*1.5*1.5)}, 1.5) {};
            \node[dot] at ({2.5 + (\startx - 2.5)*(1 + 0.18*2.8*2.8)}, 2.8) {};
            \node[dot] at ({2.5 + (\startx - 2.5)*(1 + 0.18*3.7*3.7)}, 3.7) {};
            \node[dot] at ({2.5 + (\startx - 2.5)*(1 + 0.18*4.8*4.8)}, 4.8) {};
        }
    \end{scope}
\end{scope}


\begin{scope}[shift={(12.5, 0)}]
    \node[font=\bfseries, text=black, anchor=center] at (4.5, 5.7) {Overcomplete Encoder-Decoder};
    
    \node[font=\Large] (X) at (0, 2.5) {$X$};
    
    \draw[thick, fill=blue!10, rounded corners=2pt] (1.5, 3.5) -- (3.5, 4.5) -- (3.5, 0.5) -- (1.5, 1.5) -- cycle;
    \node[font=\Large] at (2.5, 2.5) {$\phi_{enc}$};
    
    \draw[thick, fill=green!10, rounded corners=2pt] (4.2, 0.5) rectangle (4.8, 4.5);
    \node[font=\Large] (Z) at (4.5, 2.5) {$z$};
    
    \draw[thick, fill=red!10, rounded corners=2pt] (5.5, 4.5) -- (7.5, 3.5) -- (7.5, 1.5) -- (5.5, 0.5) -- cycle;
    \node[font=\Large] at (6.5, 2.5) {$\phi_{dec}$};
    
    \node[font=\Large] (Xhat) at (9, 2.5) {$\hat{X}$};
    
    \draw[->, thick] (X) -- (1.5, 2.5);
    \draw[->, thick] (3.5, 2.5) -- (4.2, 2.5);
    \draw[->, thick] (4.8, 2.5) -- (5.5, 2.5);
    \draw[->, thick] (7.5, 2.5) -- (Xhat);
    
    \node[font=\scriptsize, text=black!70] at (0, 1.8) {Low-Dim Input};
    \node[font=\scriptsize, text=black!70] at (4.5, 0.1) {High-Dim Latent Manifold};
    \node[font=\scriptsize, text=black!70] at (9, 1.8) {Low-Dim Output};
\end{scope}

\end{tikzpicture}
}
\caption{Architectural and structural mechanics overview. \textbf{Left Panel:} A topological comparison highlighting the shift from discrete sequence mapping (Residual Networks) to continuous integration (ODE Networks). The sparse vector field in the ResNet illustrates how discrete models only sample the dynamics at fixed layer intervals resulting in rigid piecewise mapping, while the dense vector field in the NODE illustrates smooth, continuous evaluation unbound by rigid depth constraints. \textbf{Right Panel:} The overcomplete architecture utilized for spatial mappings. The initial low-dimensional observation $X$ is expanded (lifted) into a high-dimensional latent state representation $z$ via the parameterized spatial encoder $\phi_{enc}$. Following continuous-time evaluations or transformations within this expanded latent space, the spatial decoder $\phi_{dec}$ projects and compresses the state back down to the reconstructed low-dimensional output $\hat{X}$.}
\label{fig:node_architecture_overview}
\end{figure*}
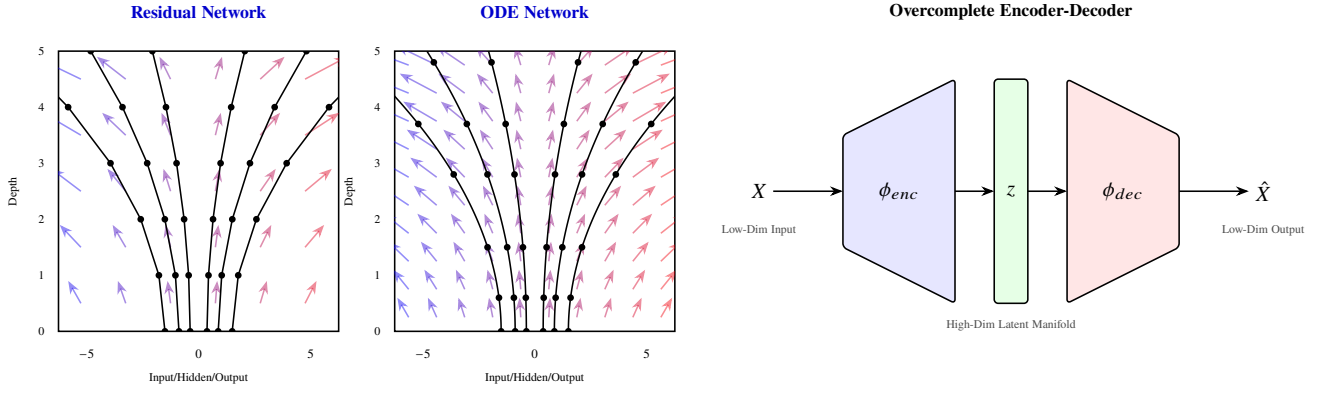

Building upon the integration of ResNets, Neural Ordinary Differential Equations (NODEs) transition the network from discrete depth to continuous depth. Rather than modeling the evolution of a hidden state $\mathbf{z}_k$ as a discrete step mapping, NODEs take the limit as $\Delta t \to 0$ and parameterize the continuous derivative of the hidden state using a neural network \cite{chen2018neural}. This can be seen in Figure \ref{fig:node_architecture_overview}. This formulates the trajectory evolution as an Initial Value Problem (IVP):
\begin{equation} 
    \frac{d\mathbf{z}(t)}{dt} = f_\theta(\mathbf{z}(t), t)
    \label{eq:node_derivative}
\end{equation}
\begin{equation} 
    \mathbf{z}(t_{k+1}) = \mathbf{z}(t_k) + \int_{t_k}^{t_{k+1}} f_\theta(\mathbf{z}(\tau), \tau) \, d\tau
    \label{eq:node_ivp}
\end{equation}

where $f_\theta$ acts as a neural surrogate vector field. For a physical observation $\mathbf{X}_0$, the standard NODE typically sets $\mathbf{z}(t_0) = \mathbf{X}_0$. The surrogate vector field $f_\theta$ is instantiated as a multi-layer perceptron (MLP) acting as a universal continuous function approximator. For an $L$-layer network, the forward propagation mapping the intermediate continuous state $\mathbf{z}(t)$ to its derivative is defined recursively as:
\begin{equation} 
    \mathbf{a}^{(0)} = \mathbf{z}(t), \quad \mathbf{a}^{(l)} = \sigma\left(\mathbf{W}^{(l)} \mathbf{a}^{(l-1)} + \mathbf{b}^{(l)}\right) \quad \text{for } l = 1, \dots, L
    \label{eq:mlp_forward}
\end{equation}
\begin{equation}
    f_\theta(\mathbf{z}(t), t) = \mathbf{W}^{(L+1)} \mathbf{a}^{(L)} + \mathbf{b}^{(L+1)}
\end{equation}

where $\mathbf{W}^{(l)}$ and $\mathbf{b}^{(l)}$ are the weight matrices and bias vectors of the $l$-th internal layer, and $\sigma(\cdot)$ represents a non-linear activation function.

\subsection{Frequency Neural ODE (FNODE) Architecture}
\label{sec:fnode_architecture}

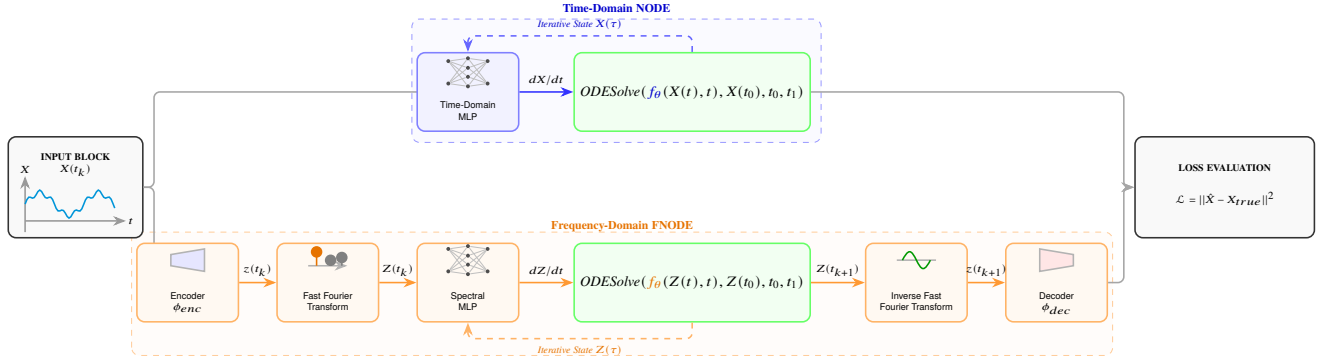
\begin{figure*}[t]
\centering
\resizebox{\textwidth}{!}{%
\begin{tikzpicture}[x=1cm, y=1cm, >=Stealth,
    stagebox/.style={rectangle, draw=black!60, thick, fill=white, rounded corners=4pt, text width=1.6cm, align=center, minimum height=1.4cm},
    solverbox/.style={rectangle, draw=green!60, thick, fill=green!5, rounded corners=4pt, text width=4.0cm, align=center, minimum height=1.4cm, font=\footnotesize},
    nodecolor/.style={fill=blue!5, draw=blue!50},
    fodecolor/.style={fill=orange!5, draw=orange!60},
    nnnode/.style={circle, fill=black!70, inner sep=0pt, minimum size=2.0pt}
]

\node[rectangle, draw=black!80, thick, fill=gray!5, rounded corners=4pt, inner sep=4pt, align=center, minimum height=1.8cm, minimum width=2.4cm] (input_box) at (-1.5, 0) {};
\node[font=\bfseries\tiny, text=black] at (-1.5, 0.55) {INPUT BLOCK};
\node[font=\tiny] at (-1.5, 0.30) {$X(t_k)$};
\begin{scope}[shift={(-2.4, -0.6)}]
    \draw[->, gray!80, thick] (0, 0) -- (1.8, 0) node[right=-2pt, font=\tiny, text=black] {$t$};
    \draw[->, gray!80, thick] (0, -0.2) -- (0, 0.8) node[above=-2pt, font=\tiny, text=black] {$X$};
    \draw[cyan!80!blue, thick, domain=0:1.6, samples=50] plot (\x, {0.2*sin(6*\x r) + 0.05*sin(30*\x r) + 0.3});
\end{scope}

\draw[->, thick, gray!70, rounded corners=3pt] (input_box.east) -- (-0.1, 0) |- (5.5, 1.7); 
\draw[->, thick, gray!70, rounded corners=3pt] (input_box.east) -- (-0.1, 0) |- (0.5, -1.7); 

\node[font=\bfseries\scriptsize, text=blue!80!black] at (8.15, 3.2) {Time-Domain NODE};

\node[stagebox, nodecolor] (n1) at (5.5, 1.7) {};
\node[below=0.05cm of n1.center, align=center, font=\tiny\sffamily] {Time-Domain\\MLP};
\begin{scope}[shift={(5.5, 2.05)}] 
    \foreach \i/\y in {1/-0.2, 2/0.2} \node[nnnode] (i\i) at (-0.35, \y) {};
    \foreach \i/\y in {1/-0.25, 2/-0.08, 3/0.08, 4/0.25} \node[nnnode] (h\i) at (0, \y) {};
    \foreach \i/\y in {1/-0.2, 2/0.2} \node[nnnode] (o\i) at (0.35, \y) {};
    \foreach \i in {1,2} \foreach \j in {1,2,3,4} \draw[gray!50, very thin] (i\i) -- (h\j);
    \foreach \i in {1,2,3,4} \foreach \j in {1,2} \draw[gray!50, very thin] (h\i) -- (o\j);
\end{scope}

\node[solverbox] (solver_top) at (9.5, 1.7) {$\mathit{ODESolve}({\color{blue!80!black}f_\theta}(X(t), t), X(t_0), t_0, t_1)$};

\draw[->, thick, blue!80] (n1.east) -- node[above, font=\tiny, text=black] {$dX/dt$} (solver_top.west);

\draw[->, thick, blue!60, dashed, rounded corners=3pt] (solver_top.north) |- (5.5, 2.7) -- (n1.north);
\node[above, text=blue!80!black, font=\tiny\itshape] at (7.5, 2.7) {Iterative State $X(\tau)$};

\begin{scope}[on background layer]
    \draw[dashed, draw=blue!40, fill=blue!2, rounded corners=4pt] (4.5, 0.8) rectangle (11.8, 3.0);
\end{scope}

\node[font=\bfseries\scriptsize, text=orange!90!black] at (8.25, -0.7) {Frequency-Domain FNODE};

\node[stagebox, fodecolor] (enc) at (0.5, -1.7) {};
\node[below=0.05cm of enc.center, align=center, font=\tiny\sffamily] {Encoder \\ $\phi_{enc}$};
\begin{scope}[shift={(0.5, -1.3)}]
    \draw[gray!60, thick, fill=blue!10] (-0.3, -0.1) -- (-0.3, 0.1) -- (0.3, 0.2) -- (0.3, -0.2) -- cycle;
\end{scope}

\node[stagebox, fodecolor] (f2) at (3.0, -1.7) {};
\node[below=0.05cm of f2.center, align=center, font=\tiny\sffamily] {Fast Fourier \\ Transform};
\begin{scope}[shift={(3.0, -1.3)}]
    \draw[gray!60, thick, ->] (-0.35, -0.1) -- (0.35, -0.1);
    \draw[orange!90!black, thick, -*] (-0.2, -0.1) -- (-0.2, 0.25);
    \draw[gray, thick, -*] (0.05, -0.1) -- (0.05, 0.1);
    \draw[gray, thick, -*] (0.25, -0.1) -- (0.25, 0.15);
\end{scope}

\node[stagebox, fodecolor] (f3) at (5.5, -1.7) {};
\node[below=0.05cm of f3.center, align=center, font=\tiny\sffamily] {Spectral\\MLP};
\begin{scope}[shift={(5.5, -1.3)}]
    \foreach \i/\y in {1/-0.2, 2/0.2} \node[nnnode] (i\i) at (-0.35, \y) {};
    \foreach \i/\y in {1/-0.25, 2/-0.08, 3/0.08, 4/0.25} \node[nnnode] (h\i) at (0, \y) {};
    \foreach \i/\y in {1/-0.2, 2/0.2} \node[nnnode] (o\i) at (0.35, \y) {};
    \foreach \i in {1,2} \foreach \j in {1,2,3,4} \draw[gray!50, very thin] (i\i) -- (h\j);
    \foreach \i in {1,2,3,4} \foreach \j in {1,2} \draw[gray!50, very thin] (h\i) -- (o\j);
\end{scope}

\node[solverbox] (solver_bot) at (9.5, -1.7) {$\mathit{ODESolve}({\color{orange!90!black}f_\theta}(Z(t), t), Z(t_0), t_0, t_1)$};

\node[stagebox, fodecolor] (f4) at (13.5, -1.7) {};
\node[below=0.05cm of f4.center, align=center, font=\tiny\sffamily] {Inverse Fast \\ Fourier Transform};
\begin{scope}[shift={(13.5, -1.3)}]
    \draw[gray!60, thick] (-0.35, 0) -- (0.35, 0);
    \draw[green!60!black, thick] (-0.25, 0) sin (-0.12, 0.15) cos (0, 0) sin (0.12, -0.15) cos (0.25, 0);
\end{scope}

\node[stagebox, fodecolor] (dec) at (16.0, -1.7) {};
\node[below=0.05cm of dec.center, align=center, font=\tiny\sffamily] {Decoder \\ $\phi_{dec}$};
\begin{scope}[shift={(16.0, -1.3)}]
    \draw[gray!60, thick, fill=red!10] (-0.3, -0.2) -- (-0.3, 0.2) -- (0.3, 0.1) -- (0.3, -0.1) -- cycle;
\end{scope}

\draw[->, thick, orange!80] (enc.east) -- node[above, font=\tiny, text=black] {$z(t_k)$} (f2.west);
\draw[->, thick, orange!80] (f2.east) -- node[above, font=\tiny, text=black] {$Z(t_k)$} (f3.west);
\draw[->, thick, orange!80] (f3.east) -- node[above, font=\tiny, text=black] {$dZ/dt$} (solver_bot.west);
\draw[->, thick, orange!80] (solver_bot.east) -- node[above, font=\tiny, text=black] {$Z(t_{k+1})$} (f4.west);
\draw[->, thick, orange!80] (f4.east) -- node[above, font=\tiny, text=black] {$z(t_{k+1})$} (dec.west);

\draw[->, thick, orange!60, dashed, rounded corners=3pt] (solver_bot.south) |- (5.5, -2.7) -- (f3.south);
\node[below, text=orange!90!black, font=\tiny\itshape] at (7.5, -2.7) {Iterative State $Z(\tau)$};

\begin{scope}[on background layer]
    \draw[dashed, draw=orange!40, fill=orange!2, rounded corners=4pt] (-0.5, -3.0) rectangle (17.0, -0.8);
\end{scope}

\begin{scope}
\node[rectangle, draw=black!80, thick, fill=gray!5, rounded corners=4pt, inner sep=6pt, align=center, minimum height=1.8cm, minimum width=3.2cm] (output_box) at (19.0, 0) {};
\node[font=\bfseries\tiny, text=black] at (19.0, 0.35) {LOSS EVALUATION};

\node[font=\tiny] at (19.0, -0.2) {$\mathcal{L} = ||\hat{X} - X_{true}||^2$};

\draw[->, thick, gray!70, rounded corners=3pt] (solver_top.east) -| (17.2, 0) -- (output_box.west);
\draw[->, thick, gray!70, rounded corners=3pt] (dec.east) -| (17.2, 0) -- (output_box.west);
\end{scope}

\end{tikzpicture}
}

\caption{How Neural ODE and Fourier Neural ODE work. The top path shows the operation of Neural ODE in the time-domain. The bottom path shows the expanded operation of Fourier Neural ODE, utilizing an encoder ($\phi_{enc}$) and decoder ($\phi_{dec}$) to map spatial observations to latent representations. The network learns purely in the frequency-domain while the physical loss is computed in the time-domain. The next continuous step for both models is iteratively computed using an $\mathit{ODE\ Solver}$ function.}
\label{fig:centered_squished_architecture}
\end{figure*}

While standard NODEs excel at modeling continuous dynamics in the spatial time domain, they can struggle to maintain phase consistency in highly oscillatory or chaotic systems. To address this, we propose the Frequency Neural ODE (FNODE). Rather than integrating the state $\mathbf{z}(t)$ directly, FNODE projects the state into the frequency domain and learns the continuous evolution of the spectral coefficients. The overcomplete encoder-decoder architecture is used to project the input from a low dimension to a higher-dimensional latent space. This ensures that the model learns the complexities of the trajectory \cite{Bourlard2022}. The encoder and decoder architecture can be seen in Figure \ref{fig:node_architecture_overview}. The difference between the NODE and FNODE architectures can be seen in Figure \ref{fig:centered_squished_architecture}.

The initial observation vector $\mathbf{X}_0 \in \mathbb{R}^{D_{data}}$, where $D_{data}$ is the dimension of the input ($\mathbb{R}^2$ for two states), is first projected into a higher-dimensional latent representation $\mathbf{z}_0 \in \mathbb{R}^{D_{spatial}}$, where $D_{spatial}$ is the specified number of latent dimensions, via a parameterized encoder $\phi_{enc}$:
\begin{equation}
    \mathbf{z}_0 = \phi_{enc}(\mathbf{X}_0)
\end{equation}

This representation is converted into the frequency domain using the discrete Fourier transform $\mathcal{F}$. To achieve spectral compression and filter high-frequency noise, a spectral mask is applied, retaining only the first $K$ frequency modes:
\begin{equation}
    \mathbf{Z}_0 = \mathcal{F}(\mathbf{z}_0)
\end{equation}
\begin{equation}
    \tilde{\mathbf{Z}}_0 = \mathbf{Z}_0[1:K]
\end{equation}

Because numerical ODE solvers are designed to operate strictly on real vector spaces, the complex spectral coefficients ($\tilde{\mathbf{Z}}$) cannot be integrated directly. They are first flattened into spectral state vector, denoted as $\mathbf{u}_0 \in \mathbb{R}^{2K}$, by concatenating their real and imaginary components:
\begin{equation}
    \mathbf{u}_0 = [\text{Re}(\tilde{\mathbf{Z}}_0) \parallel \text{Im}(\tilde{\mathbf{Z}}_0)]
\end{equation}

Following the identical continuous-depth paradigm established in Equation \ref{eq:node_derivative}, the spectral vector field is parameterized by an MLP, $f_\theta$, which uses this flattened real vector to compute the frequency derivative:
\begin{equation}
    \frac{d\mathbf{u}(t)}{dt} = f_\theta(\mathbf{u}(t), t)
\end{equation}

The continuous-time trajectory of the flattened spectral coefficients is obtained by evaluating the ODE solver from the initial time $t_0$ to the target time $t$:
\begin{equation}
    \mathbf{u}(t) = \mathbf{u}_0 + \int_{t_0}^{t} f_\theta(\mathbf{u}(\tau), \tau) \, d\tau
\end{equation}

To recover the complex spectral trajectory $\tilde{\mathbf{Z}}(t)$, the real and imaginary components outputted by the solver at time $t$ are recombined into complex numbers:
\begin{equation}
    \tilde{\mathbf{Z}}(t) = \mathbf{u}(t)_{1:K} + i \mathbf{u}(t)_{K+1:2K}
\end{equation}

The truncated modes are zero-padded back to the original uncompressed frequency length $N_{max} = (D_{spatial}/2) + 1$, utilizing a padding operator $\mathcal{P}$:

\begin{equation}
    \mathbf{Z}(t) = \mathcal{P}(\tilde{\mathbf{Z}}(t), N_{max})
\end{equation}

The continuous spatial latent state $\mathbf{z}(t)$ is then reconstructed via the inverse real Fourier transform $\mathcal{F}^{-1}$:
\begin{equation}
    \mathbf{z}(t) = \mathcal{F}^{-1}(\mathbf{Z}(t))
\end{equation}

Finally, this reconstructed latent state is projected back into the physical observation space via a parameterized decoder $\phi_{dec}$ to yield the final continuous prediction $\hat{\mathbf{X}}(t)$:
\begin{equation}
    \hat{\mathbf{X}}(t) = \phi_{dec}(\mathbf{z}(t))
\end{equation}


\section{Experimental Methodology}

To evaluate the advantages of the proposed Frequency-domain Neural Ordinary Differential Equation (FNODE) against standard continuous-depth models and discrete recurrent baselines, a comprehensive experimental framework was developed on different dynamical systems.


\subsection{Benchmark Dynamical Systems}
\label{subsec:benchmark_systems}

The models were benchmarked across four distinct dynamical systems. The systems were selected to test the robustness under varying degrees of non-linearity.

\subsubsection{The Van der Pol Oscillator}
It was originally developed in electrical engineering to model circuits with vacuum tubes, it features non-conservative, non-linear damping. The system is defined as:

\begin{equation}
    \ddot{x} - \mu(1 - x^2)\dot{x} + x = 0
\end{equation}

which translates to the first-order system:

\begin{equation}
    \begin{bmatrix} \dot{x}_1 \\ \dot{x}_2 \end{bmatrix} = \begin{bmatrix} x_2 \\ \mu(1 - x_1^2)x_2 - x_1 \end{bmatrix}
\end{equation}

where $\mu$ is a scalar that controls the non-linear damping. As $\mu$ increases, the system becomes highly \textit{stiff}. The system tends to a stable limit cycle for all initial conditions \cite{Boccara2007}. We use the value $\mu = 3.0$ in our experiment. As x approaches the maximum amplitude of the oscillation, $\dot{x}$ increases. When reaching the maximum, $\dot{x}$ rapidly switches sign and x begins to decrease slowly, building up speed in the same way as it approaches the minimum.

\subsubsection{The Duffing Oscillator}
The Duffing oscillator is a non-linear, second-order differential equation used to model damped and driven oscillators. It serves as the primary baseline for evaluating a model's ability to capture complex periodic behaviors in the frequency domain. The system is governed by:

\begin{equation}
    \ddot{x} + \delta \dot{x} + \alpha x + \beta x^3 = \gamma \cos(\omega t)
\end{equation}

where $x$ is the displacement, $\delta$ is the coefficient of damping, $\alpha$ is the linear stiffness coefficient, $\beta$ dictates the amount of non-linearity in the restoring force, and $\gamma$ and $\omega$ are the amplitude and frequency of the periodic driving force, respectively \cite{Korsch1994}. The parameter values used are $\alpha=0.5, \beta=1.0, \delta=3.0, \gamma=0.4,$ and $\omega=1.0$. For standard ODE solver compatibility, this is formulated as a system of first-order equations:

\begin{equation}
    \begin{bmatrix} \dot{x}_1 \\ \dot{x}_2 \end{bmatrix} = \begin{bmatrix} x_2 \\ \gamma \cos(\omega t) - \delta x_2 - \alpha x_1 - \beta x_1^3 \end{bmatrix}
\end{equation}

\subsubsection{The Lotka-Volterra Model}
\label{subsubsec:lotka_volterra}

This system describes the non-linear biological interaction between two competing populations. The continuous-time dynamics are governed by the following system of first-order differential equations:

\begin{equation}
    \begin{aligned}
        \dot{x} &= \alpha x - \beta x y \\
        \dot{y} &= \delta x y - \gamma y
    \end{aligned}
\end{equation}

where $x$ represents the population density of the prey, and $y$ represents the density of the predator, $\alpha$ is the prey's intrinsic exponential growth rate, $\beta$ represents the predation rate, $\delta$ denotes the predator's reproduction based on consumed prey, and $\gamma$ is the natural mortality rate of the predator in the absence of food \cite{Bacaer2011}. The parameters used in this work are $\alpha=1.0, \beta=2.0, \delta=1.0,$ and $\gamma=2.0$

\subsubsection{Lorenz Equation}
\label{subsubsec:lorenz}

It was developed by Edward Lorenz to model 2D atmospheric convection. The system is defined by a set of three coupled, non-linear ordinary differential equations:

\begin{equation}
    \begin{aligned}
        \dot{x} &= \sigma (y - x) \\
        \dot{y} &= x (\rho - z) - y \\
        \dot{z} &= x y - \beta z
    \end{aligned}
\end{equation}

where $x, y,$ and $z$ represent the system state variables, which physically correspond to the rate of convection, horizontal temperature variation, and vertical temperature variation, respectively \cite{lorenz1977}. The system's behavior is governed by three positive parameters: the Prandtl number $\sigma$, the Rayleigh number $\rho$, and a geometric factor $\beta$. We use $\sigma = 10.0, \rho = 28.0,$ and $ \beta = 8.0/3.0$. the original values used by Lorenz.

\subsection{Experimental Configuration and Hyperparameters}
\label{subsec:hyperparameters}

To ensure the reproducibility of the results and to provide a transparent basis for architectural comparison, the exact parameters governing both the data generation process and the neural network optimization are explicitly detailed below. 

\subsubsection{Curriculum Learning}
\label{sec:curriculum_learning} When a network is initialized, it typically parameterizes a highly erratic vector field. The numerical ODE solver is forced to integrate this field over a long time horizon. This introduces small localized errors that compound exponentially. \\

To overcome this, a time-horizon curriculum-learning strategy was implemented. Rather than forcing the network to predict the entire trajectory from the beginning, the model is trained progressively before attempting to train on the entire trajectory. \\

Let the complete trajectory consist of $N$ time steps spanning the interval $[t_0, t_{end}]$. The training process is divided into $K$ discrete phases. In each phase $k \in \{1, 2, \dots, K\}$, the ODE solver is strictly constrained to integrate over a truncated sequence length $L_k$, where:
\begin{equation}
    L_1 < L_2 < \dots < L_K = N
\end{equation}

During the early phases of training (e.g., $L_k = 0.1N$), a random starting index $j$ is sampled from the dataset, and the model integrates over this window $[t_j, t_{j+L_k}]$. The loss function for phase $k$ is computed exclusively over this slice:

\begin{equation}
    \mathcal{L}_k = \frac{1}{L_k} \sum_{i=j}^{j+L_k} \left\| \mathbf{y}(t_i) - \hat{\mathbf{y}}(t_i) \right\|_2^2
\end{equation}

This provides gradient stability during learning, computational traceability, and accurate learning of the dynamical system.

\subsubsection{Ensemble Learning}
\label{sec:ensemble_method}

In order to avoid local minima. As a model's convergence heavily depends on its initial parameter distribution. To mitigate these uncertainties, an ensemble approach was adopted. For every experimental configuration, an ensemble that consists of $M$ independent models (where $M=5$ in this study) is initialized and trained. \\

Each model within the ensemble, denoted as $f_{\theta_m}$ for $m \in \{1, \dots, M\}$, was initialized with a distinct random parameter state. The ensemble was optimized by minimizing the average of the individual Mean Squared Errors (MSE):

\begin{equation}
    \mathcal{L}_{ensemble} = \frac{1}{M} \sum_{m=1}^{M} \left( \frac{1}{N} \sum_{i=1}^{N} \left\| \mathbf{y}(t_i) - \hat{\mathbf{y}}_m(t_i) \right\|_2^2 \right)
\end{equation}

Because the models do not share weights and the loss components are strictly independent, this shows if the model generalizes and learns the dynamical system rather than having a "lucky" run.\\

During the evaluation phase, the variance of the ensemble provides a quantitative measurement of the variance from the models' uncertainty regarding the true physical parameters of the dynamical system. \\

For any given unseen initial condition, the predicted state distribution at time $t$ is approximated by calculating the cross-model ensemble mean $\hat{\mathbf{y}}_{\mu}(t)$ and standard deviation $\hat{\mathbf{y}}_{\sigma}(t)$:

\begin{equation}
    \hat{\mathbf{y}}_{\mu}(t) = \frac{1}{M} \sum_{m=1}^{M} \hat{\mathbf{y}}_m(t)
\end{equation}
\begin{equation}
    \hat{\mathbf{y}}_{\sigma}(t) = \sqrt{\frac{1}{M} \sum_{m=1}^{M} \left(\hat{\mathbf{y}}_m(t) - \hat{\mathbf{y}}_{\mu}(t)\right)^2}
\end{equation}

These statistics are used to generate a confidence interval around the mean prediction. This acts as a measurement for the models' generalization capability, visually distinguishing between regions where the learned physics are highly confident, which are indicated by tight uncertainty bounds.

\section{Experimental Results and Analysis}
\label{sec:results}

\subsection{Convergence Comparison}
The evaluation of the discrete and continuous-depth architectures is conducted across the four dynamical systems. To assess the generalization capabilities of each model to correctly map the trajectories, models were evaluated with the same hyperparameter values as shown in Table \ref{tab:system_meanstd}. The network depth was fixed to 2, and the width size of each layer was fixed to 32. The dataset size was fixed to 1024, which means that each model will be fed 1024 trajectories, each with a different initial condition. The $D_{spatial}$ is defined to be 128.

\begin{figure*}[]
    \centering
\includegraphics[width=1.0\textwidth]{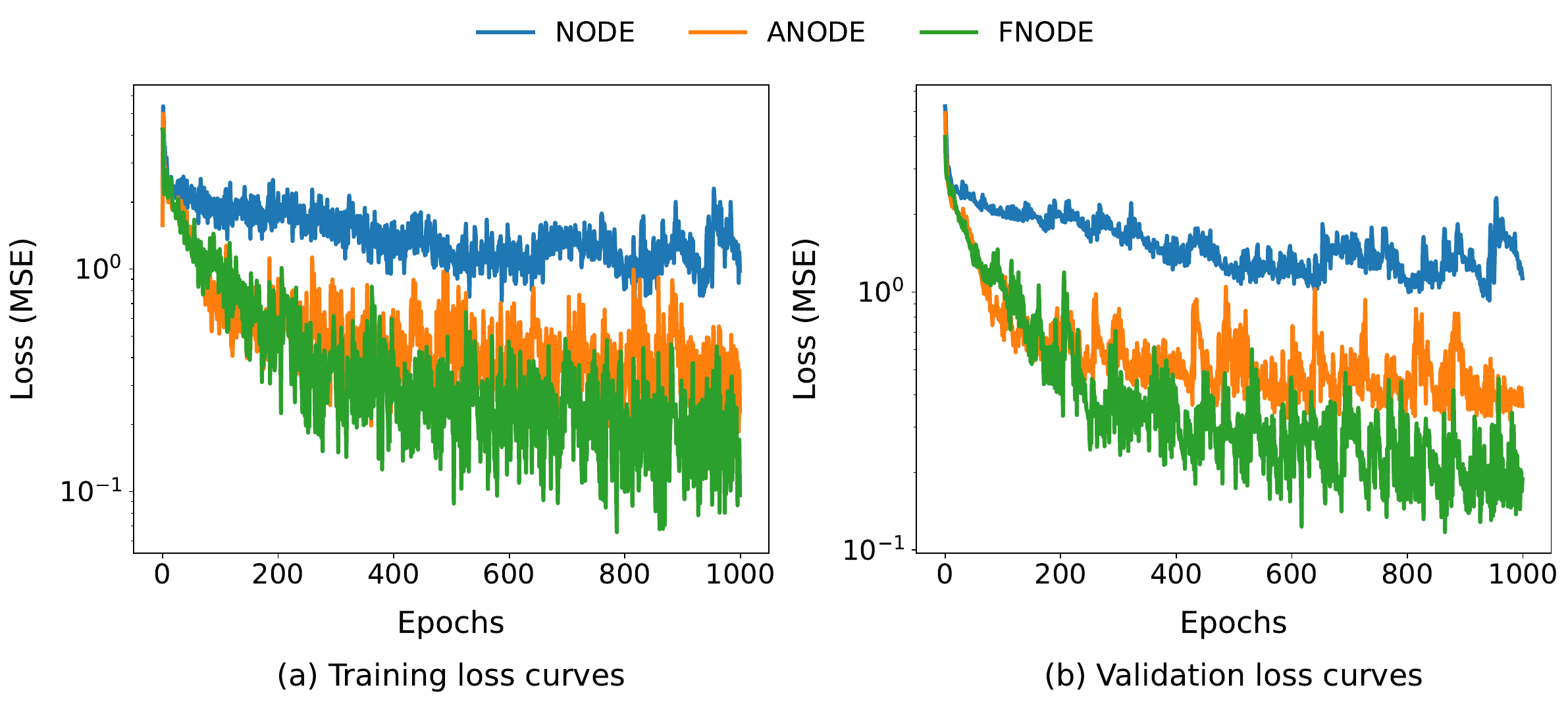}
    \caption{Loss curves for the Lotka-Volterra model for the NODE, ANODE, and FNODE}
    \label{fig:lotkavolterra_convergence}
\end{figure*}

\begin{figure*}[]
    \centering
\includegraphics[width=1.0\textwidth]{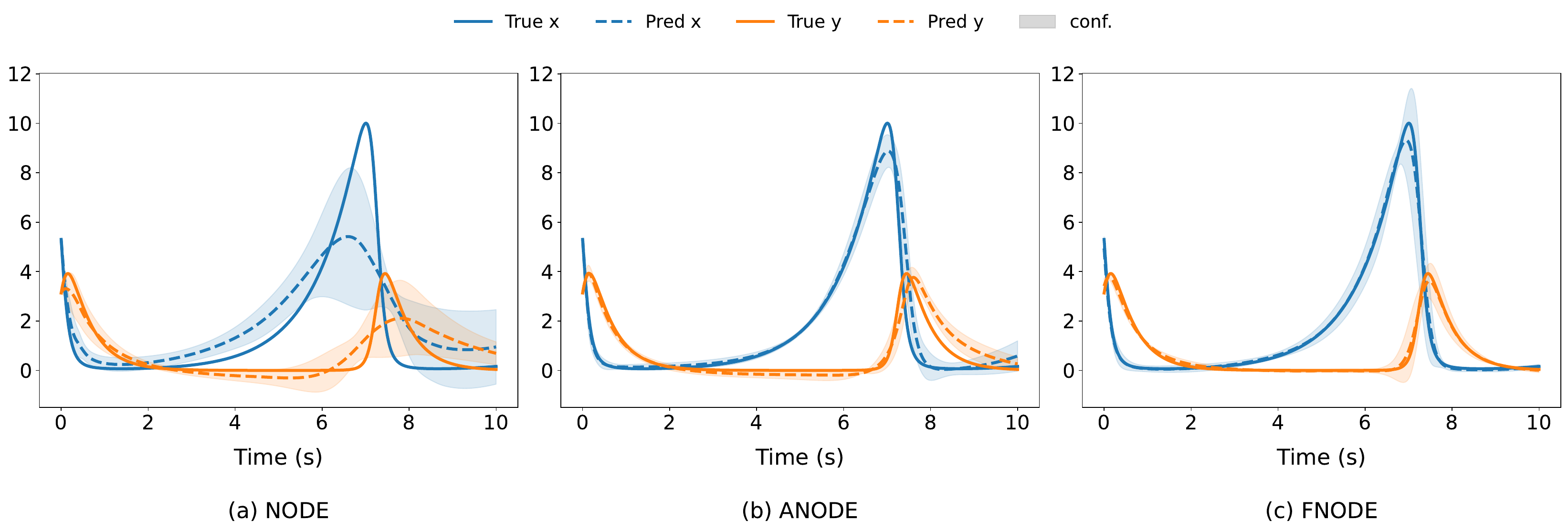}
    \caption{Prediction results for the Lotka-Volterra model for the NODE, ANODE, and FNODE}
    \label{fig:lotkavolterra_predictions}
\end{figure*}

\begin{figure*}[]
    \centering
\includegraphics[width=1.0\textwidth]{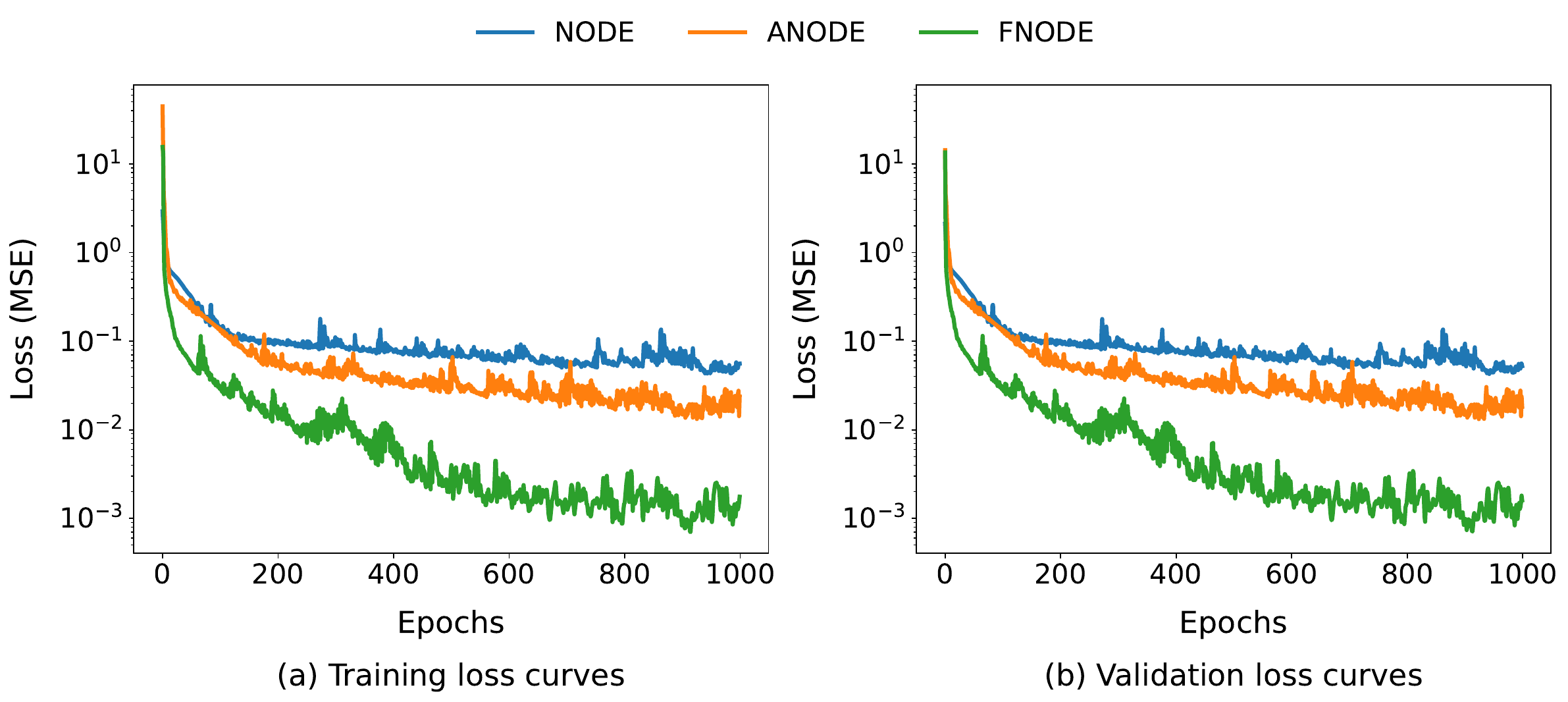} 
    \caption{Loss curves for the Duffing Oscillator model for the NODE, ANODE, and FNODE}
    \label{fig:duffing_convergence}
\end{figure*}

\begin{figure*}[]
    \centering
\includegraphics[width=1.0\textwidth]{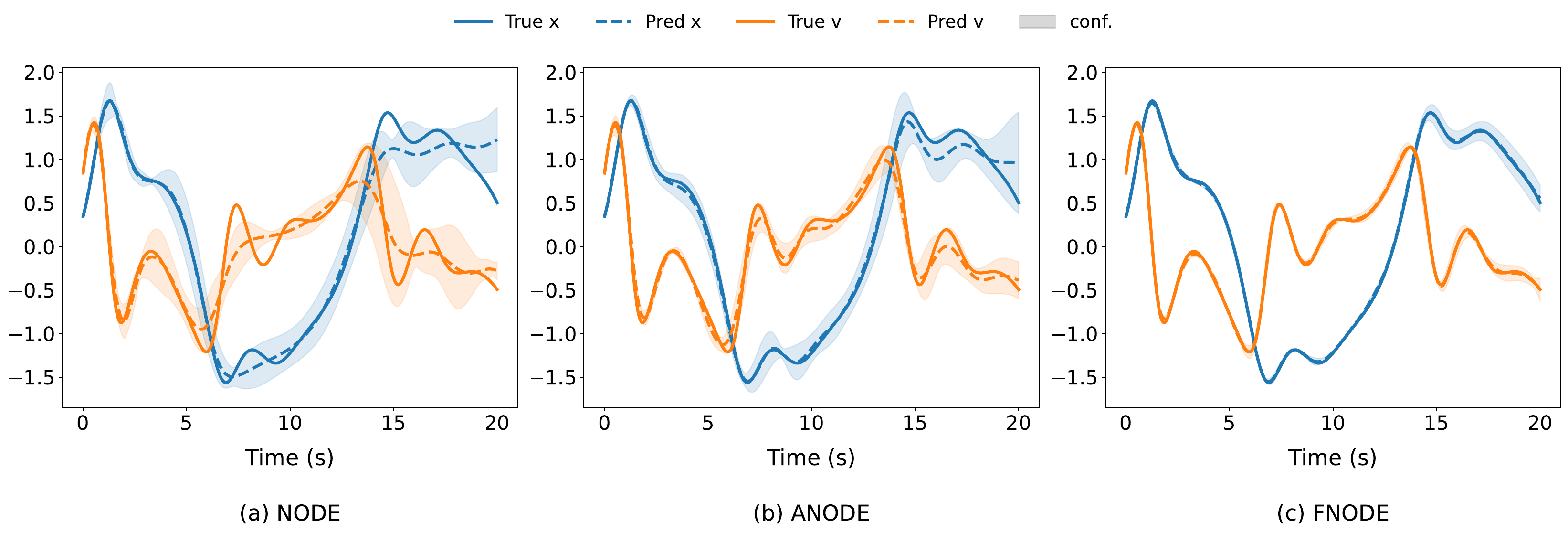} 
    \caption{Prediction results for the Duffing Oscillator model for the NODE, ANODE, and FNODE}
    \label{fig:duffing_predictions}
\end{figure*}

\begin{figure*}[]
    \centering
\includegraphics[width=1.0\textwidth]{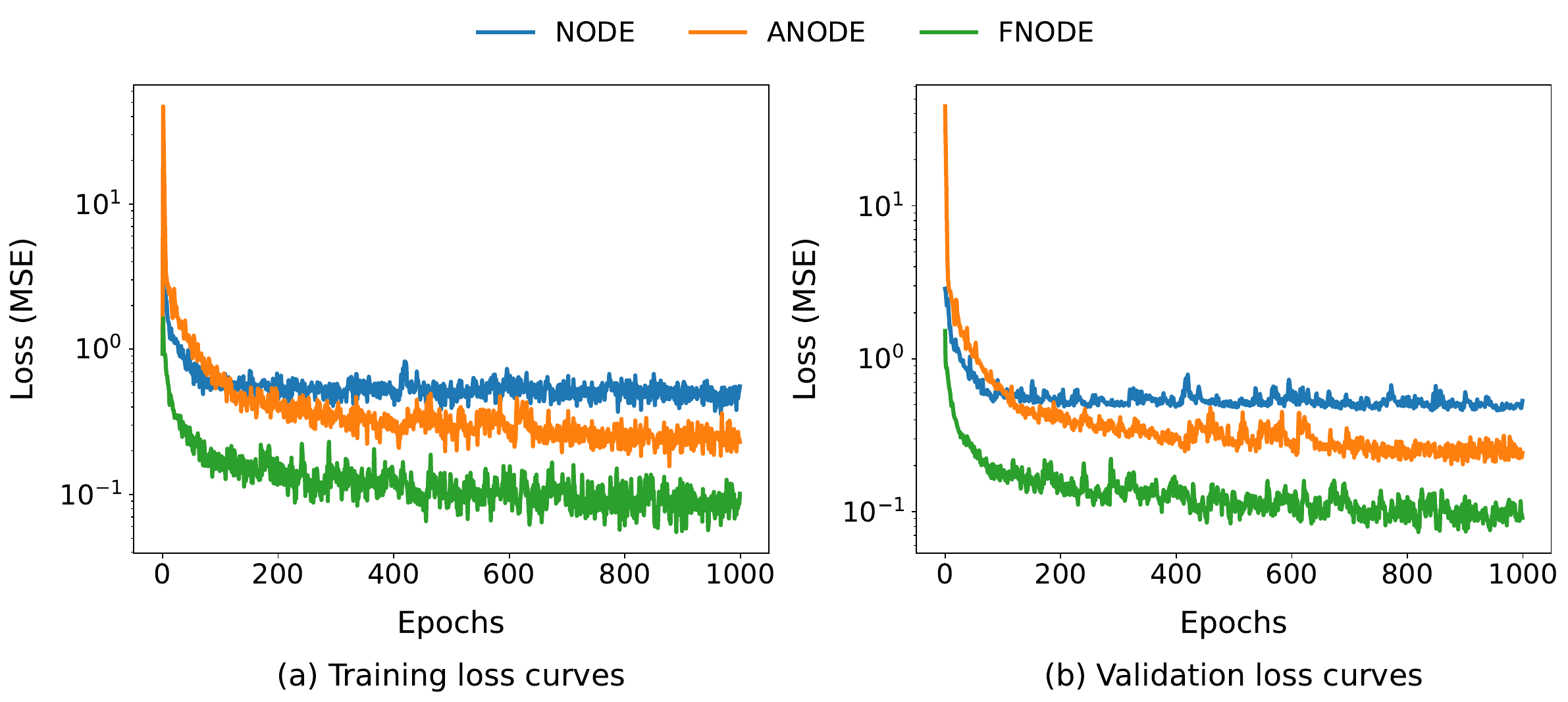} 
    \caption{Loss curves for the Van der Pol model for the NODE, ANODE, and FNODE}
    \label{fig:vanderpol_convergence}
\end{figure*}

\begin{figure*}[]
    \centering
\includegraphics[width=1.0\textwidth]{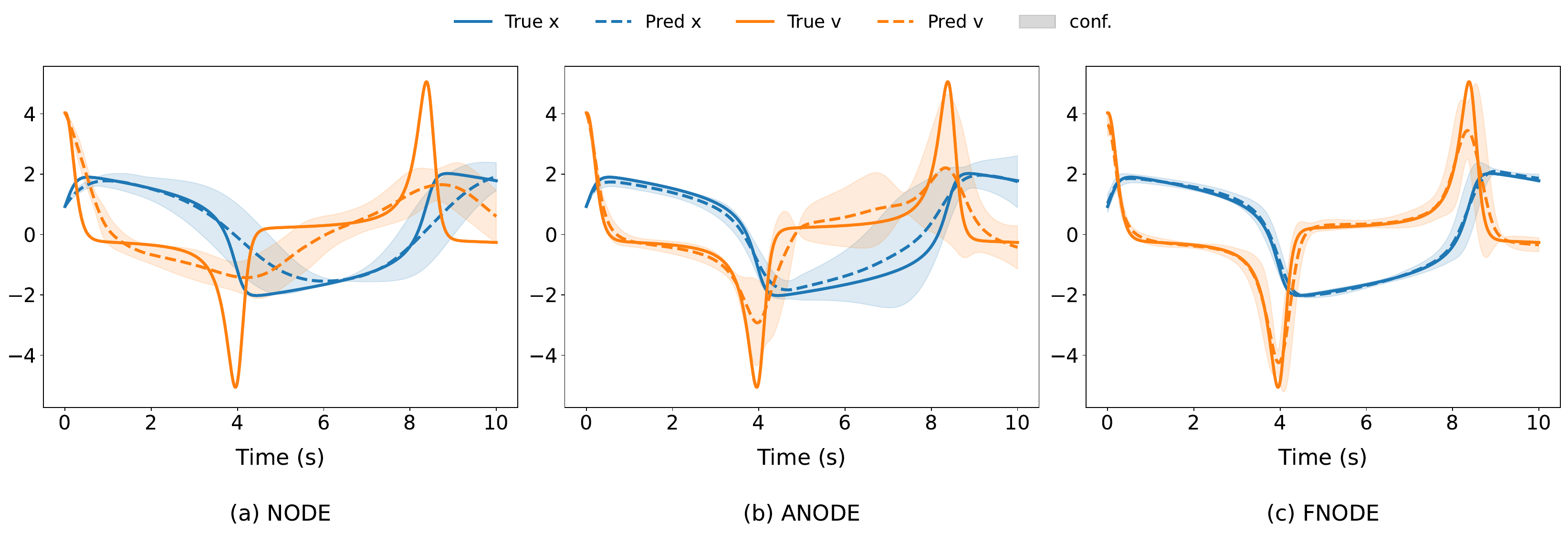} 
    \caption{Prediction results for the Van der Pol model for the NODE, ANODE, and FNODE}
    \label{fig:vanderpol_predictions}
\end{figure*}

\begin{figure*}[]
    \centering
\includegraphics[width=1.0\textwidth]{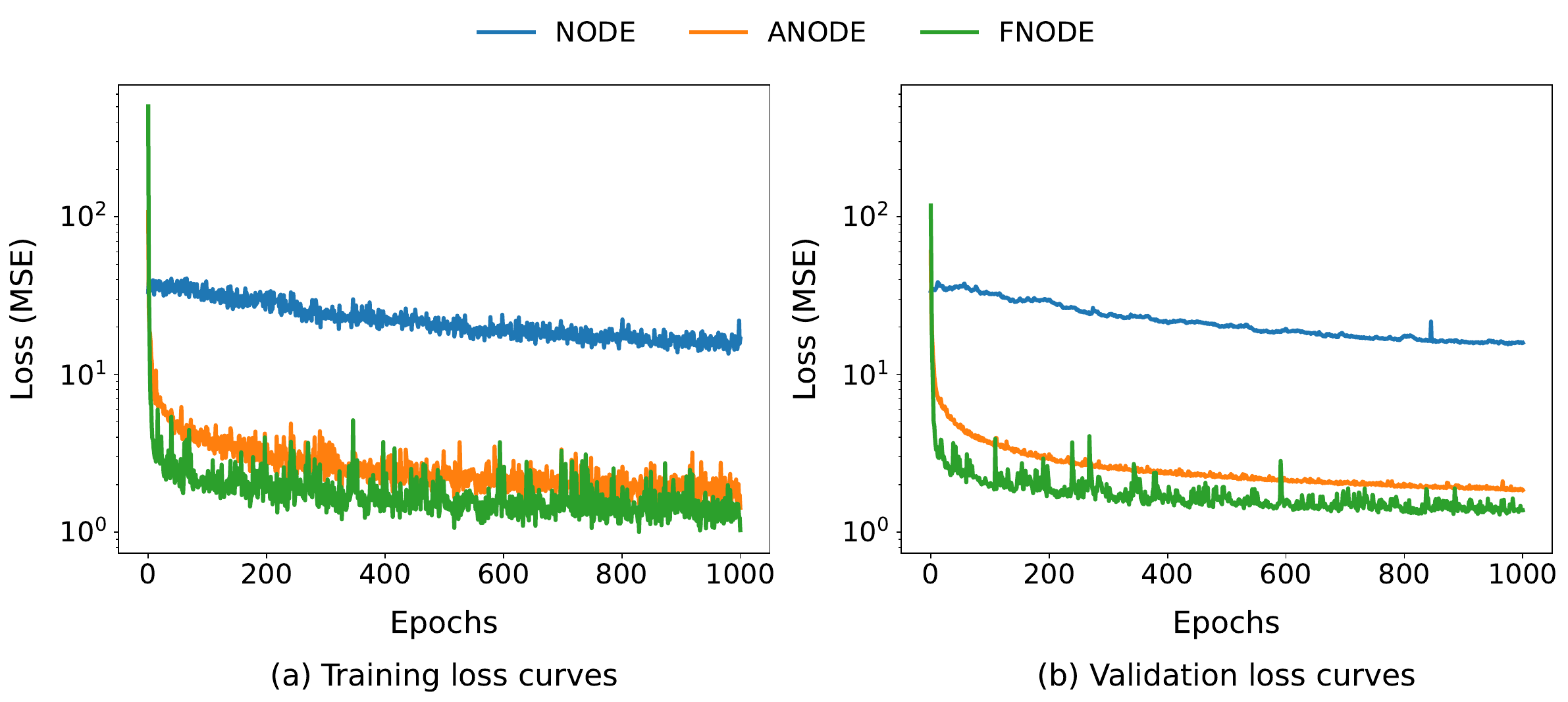} 
    \caption{Loss curves for the Lorenz model for the NODE, ANODE, and FNODE}
    \label{fig:lorenz_convergence}
\end{figure*}

\begin{figure*}[]
    \centering
\includegraphics[width=1.0\textwidth]{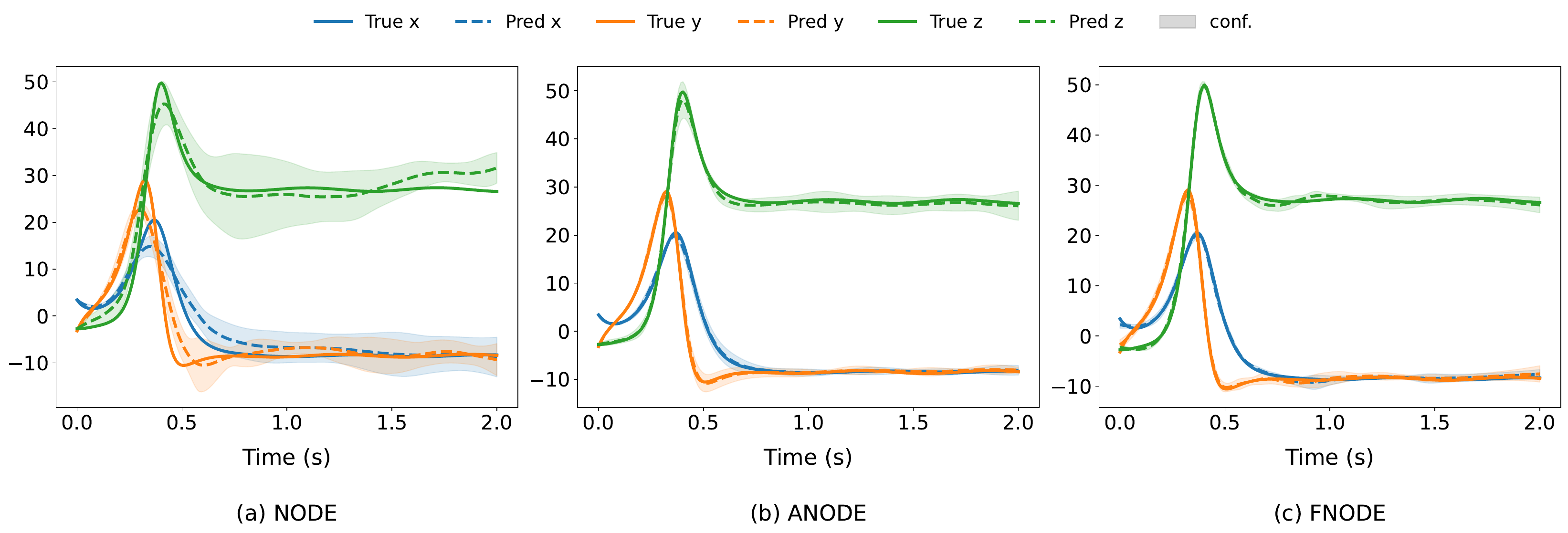} 
    \caption{Prediction results for the Lorenz model for the NODE, ANODE, and FNODE}
    \label{fig:lorenz_predictions}
\end{figure*}

The prediction results are shown from Figure \ref{fig:lotkavolterra_convergence} to \ref{fig:lorenz_predictions}. They demonstrate the performance of NODE, ANODE, and FNODE models across the four dynamical systems. The FNODE shows better convergence to the true path with a high confidence level. NODE and ANODE also show the ability to converge to the true path but with less confidence. NODE exhibited high degradation in all of the systems' predictions. This indicates that standard NODEs cannot easily map crossing trajectories. Additionally, ANDOE resolved this bottleneck by providing extra dimensions for the trajectories to untangle, improving the mean MSE. However, it seems to struggle more in the Van der Pol system, as shown in Figure \ref{fig:vanderpol_predictions}. On the other hand, FNODE achieved the lowest mean squared error (MSE) with a tight standard deviation for the test prediction, as seen in Table \ref{tab:system_meanstd}. The performance suggests that representing the dynamics as sparse Fourier coefficients provides a fundamentally more robust optimization landscape than attempting to map the temporal vector field. GRU and LSTM were also added to compare between the discrete-time and continuous-time methods. Moreover, GRU and LSTM performed better than NODE; this can be due to the fact that these models are optimized and developed for time-sequence models and can perform better than the NODE on lower hyperparameter values. Suprisingly, both GRU and LSTM performed better on the duffing oscillator than ANODE. These can be reflected back to the compounding error of the ODESolver, leading to the accumulation of prediction error.

\subsection{Ablation Study}
\label{subsec:freq_modes}

The FNODE model was tested on the dynamical systems for different frequency modes, as shown in Table \ref{tab:mape_frequency_modes}. The Median Average Percentage Error (MdAPE) was calculated for each mode. For the Duffing oscillator, the optimal number of modes is 16; after that, the percentage error starts to increase. This can be seen as the model starting to overfit, meaning that the system is no longer seen complex by the model with the current hyperparameter values. For Lotka-Volterra and Van der Pol, increasing the number of modes decreases the percentage error. However, for the Lorenz system, increasing the modes beyond 16 does not lower the MdAPE. This can be due to the fact that the model can no longer improve without changing the other hyperparameters (e.g., width size, depth).

\subsection{Limitations and Discussion}A fair assumption to be made is our model being more robust to noise due to its nature. This can be seen through the work of Wahab et al. for noise reduction and signal enhancement using Discrete Fourier Transform \cite{WAHAB2021116354}, though this needs to be verified through experimentation. We see extensions of FNODE being useful for modelling more complex systems that interact with the environment. NODE was implemented in an adaptive asynchronous control for robot systems \cite{Salehi}. \\

The first limitation of this framework is that it introduces more computational time through Fourier and inverse Fourier transforms. A study can be made to see the speed-accuracy tradeoff for the model. The second limitation is that it introduces more hyperparameters, which adds complexity during the tuning phase. Hyperparameter search algorithms may be helpful tools to determine the best combination for a system. However, this depends on whether the desired outcome is a faster or more accurate model. Overall, there are many areas of improvement in the architecture, and optimisation techniques may be adapted to modify the architecture such that it can be deployed on real-time systems. 


\section{Conclusion}
\label{sec:conclusion}

We developed the Frequency Neural Ordinary Differential Equations (FNODE). The framework was tested on four distinct dynamical systems represented by first- and second-order differential equations. Curriculum learning and ensemble methods were applied to test the generalization of models to the dynamical systems and improve gradient stability. The results have shown that the FNODE generalized better to the dynamical systems with higher confidence compared to the other models. \\

\section*{Conflict of Interest}

The authors declare that they have no known conflict of interests or personal relationships that could have appeared to influence the work reported in this paper.

\section*{Data Availability Statement}
The datasets generated and analyzed during the current study, including the synthetic time-series data for Lotka-Volterra, Duffing, Van der Pol, and the Lorenz systems, as well as the PyTorch model weights, are available from the corresponding author upon reasonable request.

\begin{nomenclature}

\EntryHeading{Letters}
\entry{$M$}{Number of Ensemble models}
\entry{$t$}{continuous time (s)}
\entry{$\Delta t$}{discrete temporal sampling interval (s)}
\entry{$\theta$}{learnable parameters (weights and biases) of the neural networks}
\entry{$\omega$}{angular frequency of the external driving force (rad s$^{-1}$)}
\entry{$x, v$}{state variables representing physical position (Duffing,Vanderpol) / prey population (Lotka-Volterra) / convection motion intensity (Lorenz)}
\entry{$y$}{state variable representing predator population density (Lotka-Volterra)/ temperature difference (Lorenz)}
\entry{$z$}{state variable representing vertical temperature profile distortion (Lorenz)}
\entry{$v$}{state variables representing physical velocity}
\entry{$\alpha$}{linear stiffness (Duffing) / prey intrinsic growth rate (Lotka-Volterra)}
\entry{$\beta$}{non-linear stiffness (Duffing) / predation rate (Lotka-Volterra) / geometric factor (Lorenz)}
\entry{$\gamma$}{driving amplitude (Duffing) / predator mortality rate (Lotka-Volterra)}
\entry{$\delta$}{damping ratio (Duffing) / predator reproduction efficiency (Lotka-Volterra)}
\entry{$\mu$}{non-linear damping parameter (Van der Pol)}
\entry{$\sigma$}{The Prandtl number (Lorenz)}
\entry{$\rho$}{the Rayleigh number (Lorenz)}

\EntryHeading{Abbreviations}
\entry{NN}{Neural Networks}
\entry{DNN}{Deep Neural Networks}
\entry{RNN}{Recurrent Neural Network}
\entry{GRU}{Gated Reccurent Unit}
\entry{LSTM}{Long Short-Term Memory Network}
\entry{MSE}{Mean Squared Error}
\entry{MdAPE}{Median Average Percentage Error}
\entry{RK4}{Runge-Kutta 4th Order Integration Method}
\entry{Tsit5}{Tsitouras 5/4, highly efficient and explicit Runge-Kutta numerical integrator
}
\entry{NODE}{Neural Ordinary Differential Equation}
\entry{ANODE}{Augmented Neural Ordinary Differential Equation}
\entry{FFT}{Fast Fourier Transform}
\entry{IFFT}{Inverse Fast Fourier Transform}
\entry{FNODE}{Frequency-domain Neural Ordinary Differential Equation}

\end{nomenclature}



\appendix   







\bibliographystyle{FNODE}   

\bibliography{FNODE_bib} 


\begin{table*}[htbp]
    \centering
    \caption{Mean and Standard Deviation results for the loss between the true and predicted paths on the test set for the five models across the three dynamical systems}
    \label{tab:system_meanstd}
    \begin{tabular}{lcccc}
        \toprule
        \textbf{Architecture} & \textbf{Lotka-Volterra} & \textbf{Duffing} & \textbf{Van der Pol} & \textbf{Lorenz} \\
        \midrule
        GRU & 1.074 ± 2.315 & 0.002 ± 0.003 & 0.143 ± 0.362 & 1.020 ± 1.940\\
        LSTM & 0.397 ± 1.03 & 0.005 ± 0.01 & 0.687 ± 1.480 & 1.607 ± 3.822\\
        NODE & 3.233 ± 7.430 & 0.034 ± 0.048 & 0.845 ± 1.348 & 5.152 ± 6.161\\
        ANODE & 0.180 ± 0.545 & 0.009 ± 0.014 & 0.270 ± 0.610 & 0.185 ± 0.260\\ 
        \midrule
        FNODE & \textbf{0.0189 ± 0.067} & \textbf{0.0003 ± 0.0002} & \textbf{0.064 ± 0.189} & \textbf{0.169 ± 0.143}\\ 
        \bottomrule
    \end{tabular}
\end{table*}

\begin{table*}[htbp]
    \centering
    \caption{Ablation Study: Median Absolute Percentage Error (MdAPE) across different FNODE Spectral Modes}
    \label{tab:mape_frequency_modes}
    \begin{tabular}{lccc}
        \toprule
        \textbf{Dynamical System} & \textbf{8 Modes} & \textbf{16 Modes} & \textbf{24 Modes} \\
        \midrule
        Duffing Oscillator   & 2.13\% & 1.83\% & 3.44\% \\
        Van der Pol          & 6.88\% & 5.94\% & 4.51\% \\
        Lotka-Volterra       & 16.6\% & 14.89\% & 13.3\% \\
        Lorenz Attractor     & 2.91\% & 2.80\% & 2.80\% \\
        \bottomrule
    \end{tabular}
\end{table*}

\end{document}